\documentclass[10pt,twocolumn,letterpaper]{article}
\usepackage[accsupp]{axessibility}  

\usepackage{cvpr}              

\usepackage{graphicx}
\usepackage{amsmath}
\usepackage{amssymb}
\usepackage{multirow}
\usepackage{multicol}
\usepackage{booktabs}

\usepackage[T1]{fontenc}

\usepackage{algorithm}
\usepackage{algorithmicx}
\usepackage{algpseudocode}

\usepackage{booktabs}

\usepackage[pagebackref,breaklinks,colorlinks]{hyperref}

\usepackage[capitalize]{cleveref}
\crefname{section}{Sec.}{Secs.}
\Crefname{section}{Section}{Sections}
\Crefname{table}{Table}{Tables}
\crefname{table}{Tab.}{Tabs.}

\def\confName{CVPR}
\def\confYear{2022}

\newcommand{\vx}{{\bf x}}

\newcommand{\vw}{{\bf w}}

\begin{document}

\title{Subspace Adversarial Training}

\author{Tao Li
\quad
Yingwen Wu
\quad
Sizhe Chen
\quad
Kun Fang
\quad
Xiaolin Huang
\and
Department of Automation, Shanghai Jiao Tong University\\
{\tt\small \{li.tao, yingwen\_wu, sizhe.chen, fanghenshao, xiaolinhuang\}@sjtu.edu.cn}
}
\maketitle

\begin{abstract}
      Single-step adversarial training (AT) has received wide attention as it proved to be both efficient and robust. However, a serious problem of catastrophic overfitting exists, i.e., the robust accuracy against projected gradient descent (PGD) attack suddenly drops to $0\%$ during the training. In this paper, we approach this problem from a novel perspective of optimization and firstly reveal the close link between the fast-growing gradient of each sample and overfitting, which can also be applied to understand robust overfitting in multi-step AT. To control the growth of the gradient, we propose a new AT method, \textbf{Sub}space \textbf{A}dversarial \textbf{T}raining (\textbf{Sub-AT}), which constrains AT in a carefully extracted subspace. It successfully resolves both kinds of overfitting and significantly boosts the robustness. In subspace, we also allow single-step AT with larger steps and larger radius, further improving the robustness performance. As a result, we achieve state-of-the-art single-step AT performance. Without any regularization term, our single-step AT can reach over $\mathbf{51}\%$ robust accuracy against strong PGD-50 attack of radius $8/255$ on CIFAR-10, reaching a competitive performance against standard multi-step PGD-10 AT with huge computational advantages. The code is released at \url{https://github.com/nblt/Sub-AT}.

\end{abstract}

\section{Introduction}
\label{sec:intro}

\begin{figure}[t]
    \centering
    \includegraphics[width=0.8\linewidth]{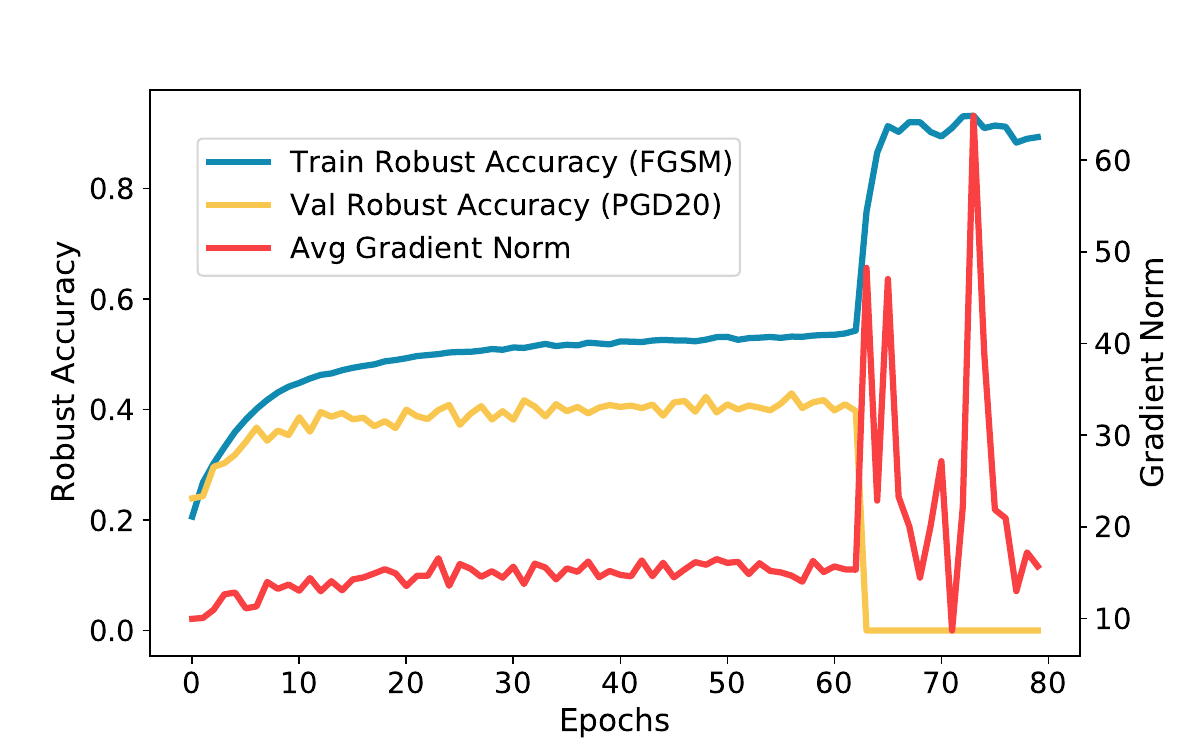}
    \caption{\emph{Catastrophic overfitting} in single-step AT. The experiments are conducted on CIFAR-10 with PreAct ResNet18 model for adversarial robustness against $\ell_\infty$ perturbations of radius $8/255$. The robust accuracy of single-step Fast AT on the validation set against PGD-20 attack abruptly drops to 0 in one single epoch, characterized by a rapid explosion of the average gradient norm of each sample. }
    \label{fig:catastropic_overfitting}
\end{figure}

Adversarial training (AT) \cite{madry2017towards}, which aims to minimize the model's risk under the worst-case perturbations,
is currently the most effective approach for improving the robustness of deep neural networks. 
For a given neural network $f(\vx, \vw)$ with parameters $\vw$, the optimization objective of AT can be formulated as follows:
\[
\min_\mathbf{w} \mathbb{E}_{(\mathbf{x},y)\sim \mathcal{D}} \left[ \max_{\delta \in \mathcal{B}(\mathbf{x}, \epsilon)} \mathcal{L} \left(f(\mathbf{x}+\delta, \mathbf{w}),y\right) \right],
\]
where $\mathcal{B}(\mathbf{x}, \epsilon)$ is the norm ball with radius $\epsilon$ and $\mathcal{L}$ is the loss function.
The key issue of AT lies in solving the inner worst-case problem by generating adversarial examples.  Presently the most efficient way to generate adversarial examples is the fast gradient sign method (FGSM)  \cite{goodfellow2014explaining}, i.e., 
\[
\mathbf{x}^{\rm adv} = \mathbf{x} + \epsilon \cdot \mathrm{sgn} \left(\nabla_\mathbf{x} \mathcal{L} (f(\vx,\vw), y)\right).
\]
Since the adversarial examples above are generated by one-step gradient propagation, the corresponding AT is called \emph{single-step} AT. 
In \cref{fig:catastropic_overfitting}, we demonstrate a standard single-step AT process where the training robust accuracy against FGSM attack keeps increasing. However, the generalization capability, i.e., the robust accuracy on the validation set under projected gradient descent (PGD) attack \cite{madry2017towards}, can suddenly drop to zero, which is a typical overfitting phenomenon referred as \emph{catastrophic overfitting} \cite{wong2020fast}.

Many works \cite{vivek2019regularizer, wong2020fast, andriushchenko2020understanding, sriramanan2020guided, kim2020understanding, kang2021understanding} are devoted to resolving such an intriguing overfitting problem.
One approach to tackle the overfitting is to use a judiciously designed learning rate schedule as well as appropriate regularizations.
For example, Wong \etal \cite{wong2020fast} proposed to add a random step to FGSM and introduce cyclic learning rates \cite{smith2017cyclical} to overcome the overfitting.
Andriushchenko \etal \cite{andriushchenko2020understanding} proposed a novel regularization term called GradAlign to further improve the quality of single-step AT solutions.
However, these methods highly rely on specifically designed learning rate schedules, 
which need to be tuned carefully for different tasks.
Another approach is to generate more precise adversarial examples. For example, Kim \etal \cite{kim2020understanding} suggested verifying the inner interval along the adversarial direction and searching for appropriate step size.
PGD AT, a typical \emph{multi-step} AT which generates adversarial examples using multiple iterations, can also help avoid catastrophic overfitting. However, these methods require multiple forward propagations. More seriously, overfitting can still prominently occur in multi-step AT (known as \emph{robust overfitting}) as demonstrated by Rice \etal \cite{rice2020overfitting}.

In order to understand this interesting phenomenon,
let us investigate what happens at the $64$-th epoch in  \cref{fig:catastropic_overfitting} when catastrophic overfitting occurs. 
Before the overfitting, 
the training robust accuracy has already stepped into a stable stage,
indicating the small norm of batch gradient $ \left \| \frac{1}{n} \sum_{i=1}^n \nabla_{\vw} \mathcal{L} (f(\vx_i^{\rm adv},\vw), y) \right \|_{2}$ ($n$ denotes the batch size). 
There are two possibilities for the small batch gradient: 
\emph{\textbf{i}}) the gradient of each sample is small; 
\emph{\textbf{ii}}) the gradients of samples does not converge, but they cancel each other, resulting in an overall balanced state. 
We then plot the average norm of each sample's gradient on one fixed training batch (i.e.,
$ \frac{1}{n}  \sum_{i=1}^n \left \|   \nabla_{\vw} \mathcal{L} (f(\vx_i^{\rm adv},\vw), y) \right \|_{2}$) in red.
An interesting thing is that before the overfitting, the average norm stays almost constant. However, it abruptly increases in the moment when the overfitting occurs.
Intuitively, at that time, the balance of gradient is broken --- the network tries to capture each sample's label with huge fluctuations, namely large gradients, a significant signal of overfitting. This phenomenon also coincides with the recent discussion on the connection between the gradient variance and generalization capability \cite{hardt2016train, neyshabur2017exploring, jiang2019fantastic}. 

\begin{figure}[htbp]
 \centering
 \begin{subfigure}{0.8\linewidth}
 \centering
  \includegraphics[width=1\linewidth]{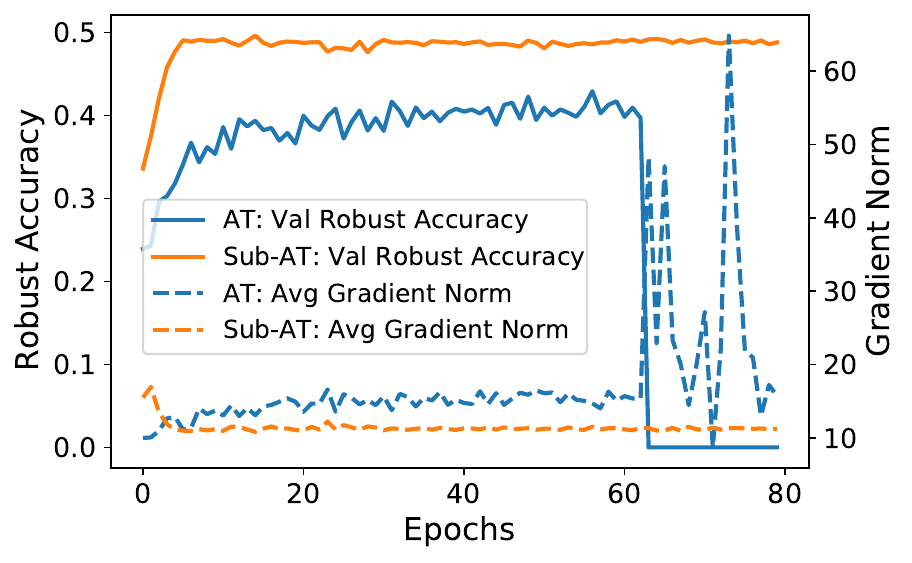}
    \caption{Single-step AT.}
    \label{singleAT}
 \end{subfigure}
 \begin{subfigure}{0.8\linewidth}
 \centering
  \includegraphics[width=1\linewidth]{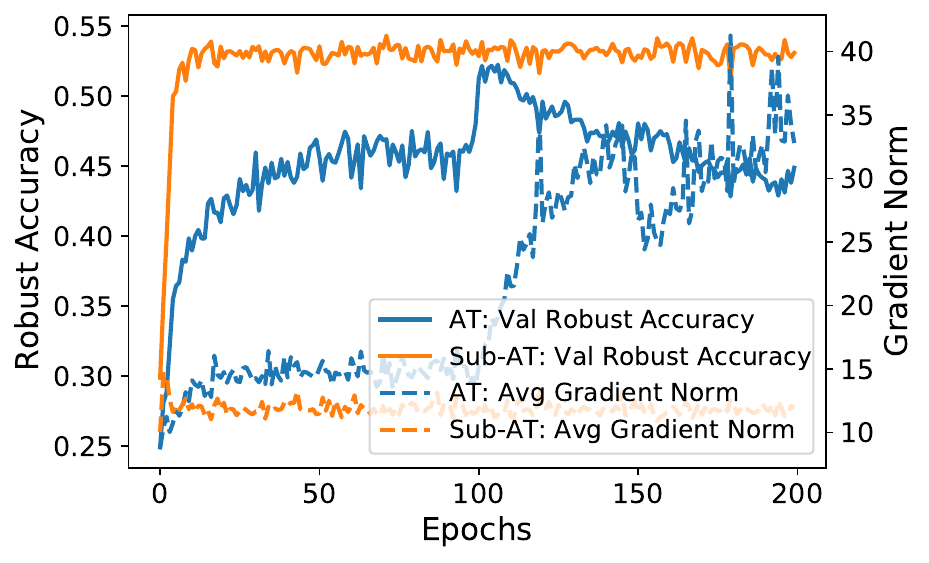}
    \caption{Multi-step AT.}
    \label{pgdAT}
 \end{subfigure}
 \caption{Resolving \emph{catastrophic overfitting} in single-step AT and \emph{robust overfitting} in multi-step AT. 
 The experiments are conducted on CIFAR-10 with PreAct ResNet18 model for robustness against $\ell_\infty$ perturbations of radius $8/255$. 
 In both overfittings, 
 robust accuracy on the validation set degenerates along with an abrupt increase of the average grad norm of the sample. Our Sub-AT successfully controls the rapid increase of the average grad norm, thereby resolving both two overfittings uniformly and significantly improving the robustness.}
 \label{fig:grad_norm_curve}
\end{figure}

Inspired by the link between large gradients and overfitting, we propose to resolve the overfitting by controlling the magnitude of the gradient.
A possible way is to restrict the gradient descent in a subspace instead of the whole parameter space,  to prevent the excessive growth of the gradient. 
The key challenge lies in keeping the network's capability in such a subspace, which has been recently discussed in \cite{li2021low} showing that, optimizing parameters in a tiny subspace extracted from training dynamics could keep the performance.
Based on this discovery, we propose a new AT method called \emph{\textbf{Sub}space \textbf{A}dversarial \textbf{T}raining} (\emph{\textbf{Sub-AT}}), which identifies such an effective subspace and conducts AT in it. 
From the training statistics of Sub-AT in \cref{singleAT},
we observe that it successfully controls the average gradient norm under a low level (the yellow dotted curve), thus resolving the catastrophic overfitting.
Meanwhile, the robust accuracy is significantly improved from 0.4 to nearly 0.5 (the yellow solid curve).
The sensitivity to learning rates is also fundamentally overcome as we only use a constant learning rate, and the results remain similar across a wide range of choices.
As a direct extension, Sub-AT can be applied to mitigate the robust overfitting (\cref{pgdAT}) in multi-step AT, implying the similar essence behind these two phenomena.
Thus for the first time, the two overfittings, which were previously 
treated separately \cite{andriushchenko2020understanding}, are now connected and resolved in a unified approach.

Since training in subspace controls the gradient magnitude and hence fundamentally resolves the catastrophic overfitting, we now can allow larger steps and radius, which previously requires the assistance of delicate regularizations, e.g. GradAlign \cite{andriushchenko2020understanding}.
It brings further improvement on robustness, from which it follows that 
pure single-step-based AT (without regularization terms) achieves competitive robustness with standard multi-step PGD AT with great computational benefits, answering a long-existing question:
\vskip 0.1cm
\noindent
\emph{Can single-step AT achieve comparable robustness against iterative attacks than multi-step AT?}
\vskip 0.1cm

\noindent
Our Sub-AT uncovers the long-neglected potential of single-step AT and can enlighten more efficient and powerful AT algorithms.

Our main contributions can be summarized as follows:
\begin{itemize}
    \item We approach the \emph{catastrophic overfitting} in single-step AT from a novel view of optimization and firstly reveal the close link between the fast-growing gradient of each sample and overfitting,
    which can also be applied to explain the \emph{robust overfitting} in multi-step AT.
    \item We propose an efficient AT method, Sub-AT, which constrains AT in a carefully extracted subspace, to control the growth of gradient.
    It uniformly resolves both kinds of overfitting, significantly improves the robustness, and successfully overcomes the sensitivity to learning rates. It is also very easy to combine with other AT methods to bring consistent improvements.
    \item Our Sub-AT achieves \emph{state-of-the-art} adversarial robustness on single-step AT and can successfully train with larger steps and larger radius, which brings further improvements. Notably, our pure single-step AT achieves over $51\%$ robust accuracy against PGD-50 attack of $\epsilon=8/255$ on CIFAR-10, competitive to the multi-step PGD-10 AT with 
    great time benefits.
\end{itemize}

\section{Related Work}
\paragraph{Adversarial Training.} 
Since deep neural networks are easily fooled by adversarial examples, many defense methods \cite{papernot2016distillation,madry2017towards,tramer2017ensemble,liao2018defense,zhang2018defense,wong2018provable,cohen2019certified,dong2018boosting,SongKNEK18,moosavi2019robustness,zhang2019you,zhang2019theoretically,wang2019improving,zhang2020attacks,croce2020robustbench,wu2020adversarial} have been proposed. Among them, AT \cite{madry2017towards}, which augments the training data with adversarial perturbations, is currently the most effective way to improve the robustness of the model. According to the number of times the gradient propagation involved in generating adversarial perturbations, AT can be mainly divided into two classes: single-step AT \cite{shafahi2019adversarial,wong2020fast,andriushchenko2020understanding,kim2020understanding, sriramanan2020guided} and multi-step AT \cite{madry2017towards, zhang2019theoretically, wang2019improving}. Single-step AT has proven to be both efficient and robust \cite{shafahi2019adversarial,wong2020fast} and thus receives much attention. For example, Free AT \cite{shafahi2019adversarial} achieves remarkable robustness performance using a single-step gradient with redundant batches and accumulative perturbations. Multi-step AT, such as PGD AT \cite{madry2017towards} and TRADES \cite{zhang2019theoretically}, generally provides better robustness guarantees than single-step AT as it generates strong adversarial perturbations. 
However, its computational cost is relatively high as multiple forward and back propagations are required during batch training.


\paragraph{Overfitting in Adversarial Training.} 
Both single-step AT and multi-step AT suffer from overfitting problems (known as catastrophic overfitting \cite{wong2020fast} and robust overfitting \cite{rice2020overfitting}, respectively) where the robust test accuracy suddenly begins to decrease as the training proceeds. The problem can be more severe in single-step AT as the robust test accuracy against PGD attack can abruptly drop to $0\%$ only in one epoch. Many works are devoted to resolving such an intriguing overfitting problem. Among them, Wong \etal \cite{wong2020fast} first suggested adding a random step to FGSM and adopting cyclic learning rates, which provides competitive robustness against PGD AT with significant time advantages. Andriushchenko \etal \cite{andriushchenko2020understanding} designed a novel regularization term called GradAlign to improve the gradient alignment inside the perturbation set and provide better robustness.
Sriramanan \etal \cite{sriramanan2020guided} introduced a relaxation term to find more suitable gradient-directions for attack.
However, these methods rely on a judiciously selected learning rate schedule, or a proper regularization coefficient \cite{kim2020understanding}. 
Towards understanding the overfitting,
Vivek \etal \cite{vivek2020single} discovered that models trained via single-step AT learn to prevent the model from generating effective adversarial examples and introduced dropout scheduling to mitigate it.
Kim \etal \cite{kim2020understanding} observed the distortion of the sample-wise decision boundary during the overfitting and suggested verifying the adversarial examples along the adversarial direction. 
However, their explanations are limited to single-step AT, and the robustness performance is still inferior. In this work, we understand overfitting from a general perspective of optimization and explain both kinds of overfitting in AT uniformly,  bringing a huge improvement in robustness.

\paragraph{Training in Subspace.}
Many works focus on the low-dimensionality essence of neural network training \cite{vinyals2012krylov, gur2018gradient, tuddenham2020quasi}.
The pioneering work \cite{li2018measuring} first proposed to train neural networks in a reduced subspace via random projection and discovered that, the required dimension to obtain $90\%$ performance of regular training is far less than the original parameters' dimension. 
The following work \cite{gressmann2020improving} improved the random bases training by considering different layers and re-drawing the random bases at each step. 
Different from random projection, Li \etal \cite{li2021low} proposed to train neural networks in low-dimensional subspaces extracted from training dynamics, and obtained comparable performance as regular training. 
We also take advantage of the subspace extracted from the training dynamics of AT and constrain the training in it, thereby successfully controlling the magnitude of the gradient and keeping the training performance.



\section{Methodology}

\subsection{Investigating Catastrophic Overfitting}
\label{investigate}
\begin{figure}
 \centering
 \begin{subfigure}{0.46\linewidth}
 \centering
  \includegraphics[width=1\linewidth]{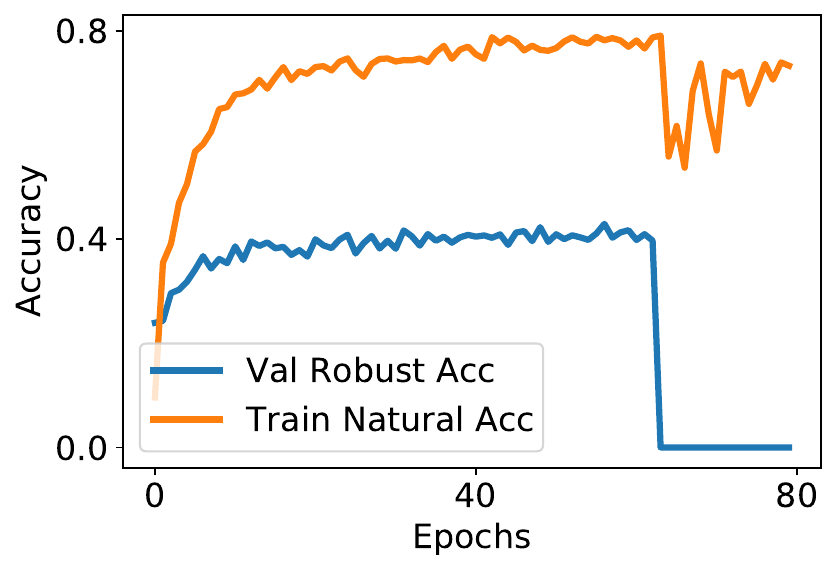}
    \caption{}
    \label{fig:overfitting_trainnatural}
 \end{subfigure}
 \begin{subfigure}{0.53\linewidth}
 \centering
  \includegraphics[width=1\linewidth]{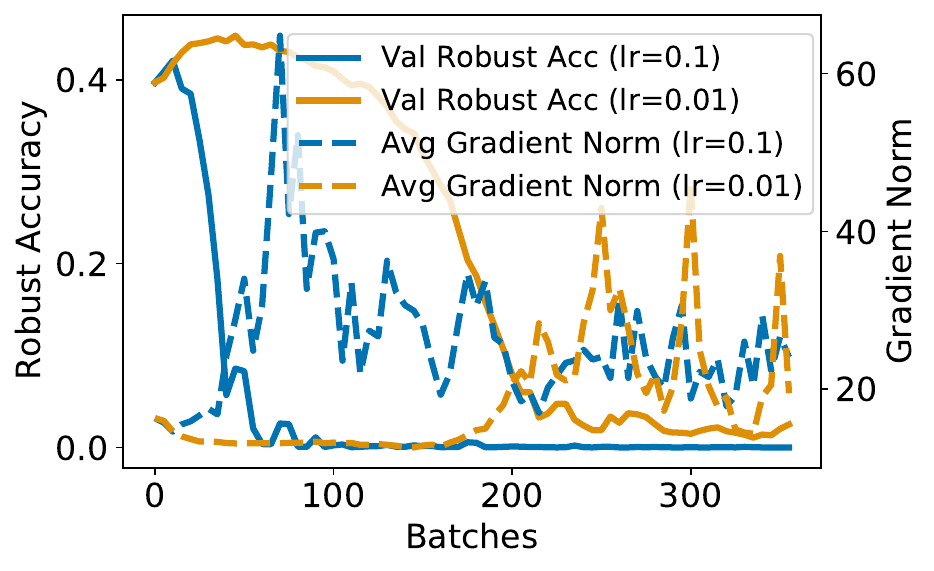}
    \caption{}
    \label{fig:investigate}
 \end{subfigure}
 \caption{ 
 (a) natural accuracy on training data also collapses at the overfitting; (b) the variation of the statistics in the 64th epoch: switching to a smaller learning rate could alleviate the catastrophic overfitting with a better-controlled gradient norm.}
 \label{fig:investigate_sum}
\end{figure}

First, we focus on an interesting phenomenon in  \cref{fig:overfitting_trainnatural}: when catastrophic overfitting occurs, the natural accuracy on training data goes through a collapse. Recall that at the same time, the robust training accuracy goes through a sudden increase, as illustrated in  \cref{fig:catastropic_overfitting}. 
These two phenomena suggest that before the overfitting, the network learns robust features that benefit both robust and natural accuracy. 
However, when overfitting occurs, the network turns to capture the adversarial information in training data (with adversarial perturbations), which harms natural accuracy, i.e., generalization capability to natural examples.
Further, it loses generalization capability to new adversarial examples generated by PGD attacks, 
as they may lie very close to natural examples.
The adversarial information is so ``hard'' to learn that, the network has to go through a huge fluctuation --- and eventually overfit it --- resulting in nearly zero robust accuracy on test data.


Since adversarial examples are relatively ``hard'' to learn,
we pay attention to the evolution of each sample's gradient and consider the average norm of the gradient to analyze the training status. Specifically, we record the robust accuracy of Fast AT \cite{wong2020fast} against PGD-20 attack and the average norm of the gradient in one fixed training batch with size $n=256$ during the training (the estimations for batch normalization \cite{ioffe2015batch} are frozen for sample-wise gradient estimation).  \cref{fig:catastropic_overfitting} illustrates the statistics when the catastrophic overfitting occurs, where we observe that the abrupt decrease of the robust accuracy highly coincides with the sudden increase of the average gradient norm. This phenomenon implies that during the catastrophic overfitting, the gradient of each sample suddenly increases, resulting in a huge fluctuation in training and, eventually, significant degeneration on robust generalization. 


To further investigate the link between increasing gradients and overfitting, we examine the detailed statistics in the 64th epoch, when catastrophic overfitting happens. We record the statistics after each iteration. For comparison, we also consider decaying the learning rate from $0.1$ to $0.01$ before the 64th epoch training, as increasing gradient indicates an excessive learning rate.
The results are illustrated in  \cref{fig:investigate}, where we observe that although for both learning rates catastrophic overfitting eventually occurs, a smaller learning rate could help. It achieves better robustness accuracy and remarkably postpones the overfitting with a better-controlled average gradient norm (the yellow dotted curve). We conclude the findings as follows: \emph{\textbf{i}}) the overfitting is closely related to the fast-growing of the average gradient norm, and  a delicately chosen learning rate could help suppress the growth of the gradient, which results in recent advances on adopting heuristic learning rates \cite{wong2020fast};
\emph{\textbf{ii}}) to control the catastrophic overfitting, we have to control the growth of the average gradient norm. 

\subsection{Controlling the Gradient Magnitude}
Let us focus on how to control the gradient magnitude of each sample during the training. Our main idea is to constrain the gradient descent of AT in a low-dimensional subspace instead of the whole parameter space, thus implicitly suppressing the fast growth of the gradient.
First, we consider how to obtain such a subspace that is effective for AT. Recently, Li \etal \cite{li2021low} proposed an algorithm called DLDR, which effectively extracts a low-dimensional subspace for optimization from the training trajectory. It generally contains two steps: 
\begin{itemize}
    \item \textbf{Step 1:} sample  model checkpoints $ \{\mathbf{w}_1,\mathbf{w}_2,\dots,\mathbf{w}_t\}$ during the regular training where we align the model parameters as a vector $\mathbf{w}_i$ with length of the parameters' number $N$; 
    \item \textbf{Step 2:} perform singular vector decomposition (SVD) on the aligned parameter matrix $ \left [ \mathbf{w}_1,\mathbf{w}_2,\dots,\mathbf{w}_t\right ]$ and obtain mutually orthogonal  bases of the subspace $[\mathbf{u}_1,\mathbf{u}_2,\dots,\mathbf{u}_d]$ with dimensionality $d$.
\end{itemize}
In this work, we apply DLDR algorithm \cite{li2021low} to extract the effective subspace for AT. 
Note that we sample the model parameters \emph{before} the overfitting occurs since in this period the network learns robust features beneficial for both robust and natural accuracy,  as demonstrated in \cref{investigate}.
We expect that optimizing the network in such an  extracted subspace could overcome the overfitting and obtain good robustness.
After extracting the subspace, we rewind the model parameters to initialization and constrain AT optimization in subspace by projecting the gradient onto it. The detailed algorithm is summarized in  \cref{Sub-AT}.

\begin{algorithm}
\caption{Subspace Adversarial Training (Sub-AT)}\label{Sub-AT}
\begin{algorithmic}[1]
\Require The dimensionality of the subspace $d$, the number of sampling times $t$ for DLDR, learning rate $\alpha$, batch size $n$ and training data $\left \{ \left (\mathbf{x}_i, y_i  \right  )\right \}$;
\vskip 0.2cm
\State \textbf{Phase 1:} obtaining the orthonormal bases of subspace $[\mathbf{u}_1,\mathbf{u}_2,\dots,\mathbf{u}_d]$;
\State Sample a parameter trajectory  $ \{\mathbf{w}_1^s,\mathbf{w}_2^s,\dots,\mathbf{w}_t^s\}$ along AT training with a certain strategy;
\State $\overline{\mathbf{w}}=\frac{1}{t}\sum_{i=1}^t \mathbf{w}_i^s$;
\State $W=[\mathbf{w}_1^s-\overline{\mathbf{w}},\mathbf{w}_2^s-\overline{\mathbf{w}},\ldots,\mathbf{w}_t^s-\overline{\mathbf{w}}]$;
\State Perform spectral decomposition on $W^{\top}W$ and obtain the largest $d$ eigenvalues $[\sigma_1^2,\sigma_2^2,\dots,\sigma_d^2]$ with corresponding eigenvectors
$[\mathbf{v}_1,\mathbf{v}_2,\dots,\mathbf{v}_d]$;
\State $ \mathbf{u}_i = \frac{1}{\sigma_i} W \mathbf{v}_i$;
\vskip 0.2cm
\State \textbf{Phase 2:} conducting AT in extracted subspaces;
\State $k \gets 0$;
\State $P = [\mathbf{u}_1,\mathbf{u}_2,\dots,\mathbf{u}_d]$;
\State $\mathbf{w}_0 = \mathbf{w}_1^s $;
\While{not converging}
\State Sample mini-batch data $\left \{ \left (\mathbf{x}_i, y  \right )  \right \}_{i=1}^n$;
\State Generate adversarial examples $\left \{ \mathbf{x}_i^{\rm adv} \right \}_{i=1}^n$;
\State $\mathbf{g}_k^{\rm adv} \gets \frac{1}{n} \sum_{i=1}^n \nabla_{\vw} \mathcal{L} (f(\vx_i^{\rm adv},\vw_k), y)$;
\State $\mathbf{w}_{k+1} \gets \mathbf{w}_k - \alpha P \left ( P^{\top} \mathbf{g}_k^{\rm adv} \right )$; 
\State $k \gets k + 1$;
\EndWhile
\State Return $\mathbf{w}_{k}$;
\end{algorithmic}
\end{algorithm}

\paragraph{Sampling Strategy.}
We adopt a simple sampling strategy for DLDR: sampling twice uniformly in each epoch training. A more delicate strategy is promising to improve the performance.
For sampling epochs, we expect that the best performance will be obtained with sampling right before the overfitting. Sampling in the start of the training is not good, 
as the subspace cannot be well estimated. We conduct DLDR with a bit more epochs in a safe region when the overfitting certainly has not happened.
The dimensionality of the subspace $d$ is set to $80$ on CIFAR-10 for single-step AT by default. We provide a detailed sampling strategy in 
\cref{sec:samplingstragegy}.

\begin{figure}
 \centering
 \begin{subfigure}{0.62\linewidth}
 \centering
  \includegraphics[width=1\linewidth]{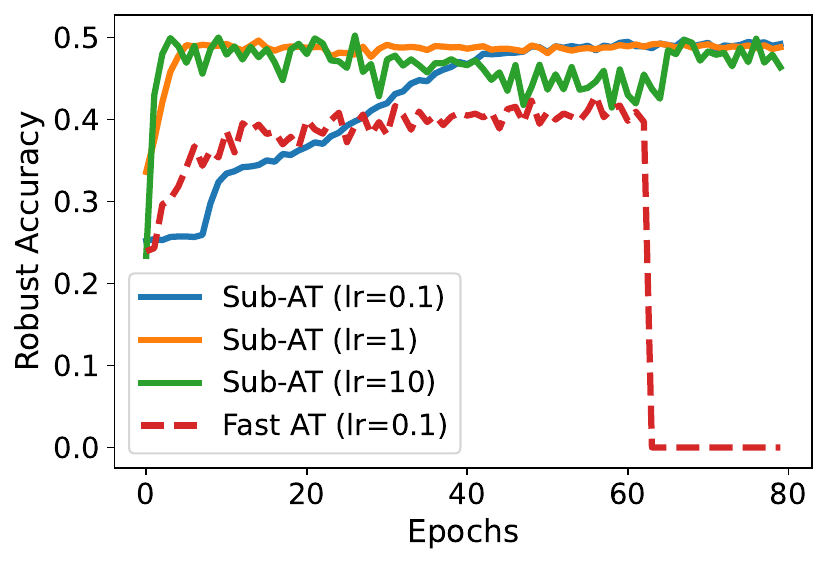}
    \caption{}
    \label{fig:lr}
 \end{subfigure}
 \begin{subfigure}{0.37\linewidth}
 \centering
  \includegraphics[width=1\linewidth]{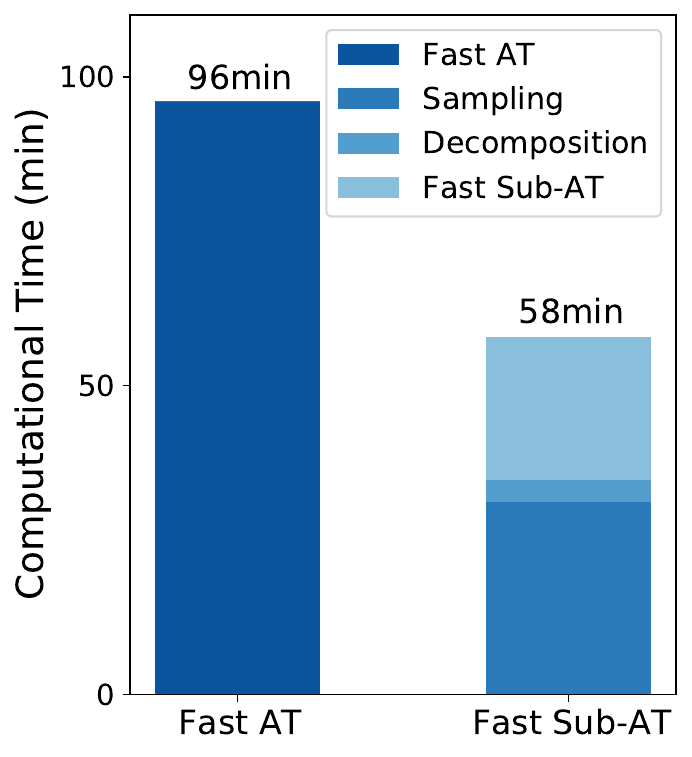}
    \caption{}
    \label{fig:timeconsumption}
 \end{subfigure}
 \caption{ (a) Sub-AT is extremely robust to learning rate; (b) detailed time consumption analysis on CIFAR-10.}
 \label{fig:investi}
\end{figure}

\paragraph{Training Performance.}
In  \cref{fig:grad_norm_curve}, we demonstrate that Sub-AT successfully controls the average gradient norm under a constant low level, thereby resolving both catastrophic and robust overfitting meanwhile significantly improving the robustness performance.
Then in  \cref{fig:lr}, we demonstrate that Sub-AT is highly robust to a wide range of learning rates and can converge only with a constant learning rate.
Notably, with a large learning rate, Sub-AT can converge quickly in a few epochs and keep the performance without overfitting.
Thus Sub-AT fundamentally overcomes the sensitivity to learning rates and obtains true robustness to overfitting. In this work, we set the constant learning rate as 1 by default for both stable and efficient training.

\paragraph{Computational Analysis.} 
The computational overhead of Sub-AT consists of two parts: DLDR and subspace training. The DLDR part contains two steps: sampling and decomposition. The time consumption on the decomposition is negligible compared to the sampling as it only involves a spectral decomposition of a $t \times t$ matrix and two matrix productions. The training of Sub-AT has almost the same computational complexity as the standard AT with little additional cost on the gradient projection. Detailed computational analysis on CIFAR-10 is illustrated in \cref{fig:timeconsumption}. We observe that Sub-AT reduces total training time overhead compared to standard AT training as it only samples a piece of the trajectory with efficient training in the subspace.

\section{Experiments}
In this section, we conduct comprehensive experiments to verify the effectiveness of Sub-AT in resolving the overfitting of both single-step and multi-step AT. We first demonstrate that Sub-AT successfully resolves catastrophic overfitting and achieves state-of-the-art robustness performance in single-step AT. Then we show that after obtaining the subspace, Sub-AT allows larger steps and larger radius, which further improves the robustness.
Finally, we apply Sub-AT to mitigate robust overfitting in multi-step AT and show that leveraging the subspace, weak single-step AT is able to achieve even better robustness than multi-step AT, revealing the great potential of single-step AT.

\begin{table*}[htbp]
 \centering
 \caption{Performance comparisons of single-step AT on CIFAR-10/100. The robustness is evaluated under \textbf{PGD-50} attack.}
 \label{tab:single-step}
 {\small
 \begin{tabular}{c|cc|cc|cc|cc|cc}
 \toprule
 \multirow{3}{*}{Schedule} &\multirow{3}{*}{Method}&\multirow{3}{*}{Subspace} &\multicolumn{4}{c|}{{CIFAR-10}} &\multicolumn{4}{c}{{CIFAR-100}}\\
 \cline{4-11}
  & & &\multicolumn{2}{c|}{Robust Accuracy} &\multicolumn{2}{c|}{Natural Accuracy} & \multicolumn{2}{c|}{Robust Accuracy} &\multicolumn{2}{c}{Natural Accuracy} \\
 \cline{4-11}
  & & &Best &Final &Best &Final &Best &Final &Best &Final \\
\midrule

 cyclic &Fast AT &-- &45.82 &45.69 &82.36 &83.26 &16.72 &0.00 &34.51 &47.23\\
 cyclic &GradAlign &-- &47.02 &46.73 &80.43 &81.34 &24.57 &24.22 &50.82 &51.92\\
\midrule
 piecewise &Fast AT &-- &39.95 &0.00 &73.13 &89.93 &17.84 &0.00 &41.54 &46.46\\
 piecewise &Single AT &-- &35.48 & 32.43 & 83.68 & 86.86 & 16.18 & 0.91 & 55.88 & 59.15\\
 piecewise &Free AT &-- &47.30 & 47.00 & 79.37 & 79.98 & 23.50 & 22.93 & 50.91 & 51.64 \\
 piecewise &GradAlign &-- &42.16 &0.02 &71.64 &88.77 &23.80 &15.60 &49.30 &53.40\\
 piecewise &GAT &-- &50.03 &41.37 &82.38 &84.45 &23.12 &20.40 &58.33  &57.09\\
\midrule
 constant &Fast Sub-AT &Fast AT &48.22 &47.88 &82.36 &82.74 &24.97 &24.55 &52.74 &53.09\\
 
 constant &GradAlign Sub-AT &GradAlign &{48.88} &{48.40} &79.82 &80.84 &\textbf{25.69} &\textbf{25.46} &52.65 &52.92\\
 constant &GAT Sub-AT &GAT &\textbf{51.15} &\textbf{50.80} &81.76 &81.61 &23.40 &22.96 &57.71 &58.45 \\
 \bottomrule
 \end{tabular}
 }
\end{table*}
\begin{table*}[htbp]
 \centering
 \caption{Results of single-step Sub-AT with a larger training $\epsilon$ against $\ell_\infty$ perturbations of radius $8/255$ on CIFAR-10 ($\alpha=1.25\epsilon$).
}
 \label{tab:largestepandlargeradius}
 {\footnotesize 
 \begin{tabular}{l|c|ccc|ccc|c}
 \toprule
    \multirow{2}{*}{Method} &\multirow{2}{*}{Subspace} &\multicolumn{3}{c|}{Best} &\multicolumn{3}{c|}{Final} &\multirow{2}{*}{Time}\\
    \cline{3-5} \cline{6-8}
     & &Natural &PGD-50 &AA &Natural &PGD-50 &AA &\\
 \midrule
  Fast AT &--  &73.13 &39.95 &37.55 &89.93 &0.00 &0.00 &1.6h\\
Fast Sub-AT ($\epsilon=8/255$)  &Fast AT   &82.36 &48.22 &44.20 &82.74 &47.88 &43.89 &1.0h\\
 Fast Sub-AT ($\epsilon=12/255$) &Fast AT &80.74 &50.38 &45.84 &80.91 &49.64 &45.40 &1.0h \\
 Fast Sub-AT ($\epsilon=16/255$) &Fast AT &78.64 &{51.46} &{46.11} &79.13 &{51.22} &{46.03} &1.0h\\
 Fast Sub-AT ($\epsilon=12/255$) &GAT &80.77 &52.41 &46.80 &80.72 &52.30 &46.80  &2.1h\\
 Fast Sub-AT ($\epsilon=14/255$) &GAT &79.96 &\textbf{53.35} &\textbf{47.25} &80.14 &\textbf{53.02} &\textbf{46.92} &2.1h\\
 \midrule
 PGD-10 AT &-- &80.50 &50.79 &47.29 &82.92 &42.51 &41.08 &7.0h\\
 \bottomrule
 \end{tabular}
 }
\end{table*}

\subsection{Experiment Setup}
\noindent
\textbf{Datasets.}
Three datasets are considered in our experiments: CIFAR-10/100 \cite{krizhevsky2009learning} and Tiny-ImageNet \cite{deng2009imagenet}. 
We randomly split the original training set as training and validation set according to a ratio of 9:1 \cite{chen2020robust}. Due to the limited space, we place the Tiny-ImageNet results in 
\cref{TinyImageNet}. 

\noindent
\textbf{Attack.}
We consider two typical types of adversarial perturbations: 
$\ell_\infty$ norm with radius $\epsilon=8/255$ and $\ell_2$ norm with radius $\epsilon=128/255$.
For single-step AT, we focus on $\ell_\infty$ norm attack and use the recommended step size of $\alpha=1.25\epsilon$ described by Wong \etal \cite{wong2020fast}.
For multi-step AT, we generate adversarial perturbations with $10$ steps attacks of step size $\alpha=2/255$ for $\ell_\infty$ norm and $\alpha=15/255$ for $\ell_2 $ norm, the standard PGD AT following the setting of Rice \etal \cite{rice2020overfitting}. 
We consider PGD-20 \cite{rice2020overfitting},
PGD-50 (with $50$ iterations and $10$ restarts) \cite{wong2020fast} and also Auto-Attack, a strong and reliable attack recently proposed by \cite{croce2020reliable},
for a rigorous evaluation on robustness.

\noindent
\textbf{Training.}
For all experiments, we use PreAct ResNet-18 \cite{he2016deep} model as a default choice. Experiments with Wide-ResNet \cite{zagoruyko2016wide} can be found in 
\cref{WideResNet}. 
Three learning rate schedules are considered: 
\textbf{\emph{i}}) \emph{cyclic} schedule \cite{wong2020fast, andriushchenko2020understanding} which can help overcome the overfitting; 
\textbf{{\emph{ii}}}) \emph{piecewise} schedule, i.e., training the model for $200$ epochs with an initial learning rate $0.1$ and decaying by ten at $100$ and $150$ epochs (as our default setting for base AT), which is commonly used and produces best robustness performance as suggested by Rice \etal \cite{rice2020overfitting}; 
\textbf{\emph{iii}}) \emph{constant} schedule, which is adopted for our Sub-AT to show its insensitivity to learning rates.
We set the learning rate as $1$ \emph{without} a schedule and train the model for $40$ epochs in subspace with a sufficient convergence.
We use a batch size of 128 and SGD optimizer with momentum 0.9 and weight decay $10^{-4}$.
Data augmentations, such as 4-pixel padding, random cropping, and horizontal flipping, are applied.

\noindent
\textbf{Evaluation.}
We consider both the best and final robustness performance during the training and use the difference between them to evaluate the degree of overfitting. The model checkpoint that achieves the best robust accuracy on the validation set is selected as the best model, while the final is an average of the last five epochs \cite{rice2020overfitting} (except that cyclic learning rates use the last epoch).
The time consumption is measured on an Nvidia Geforce GTX 2080 TI.
For Sub-AT, we repeat over five independent runs.
The standard deviations in tables are omitted as they are very small ($\leq 0.45 \%$), which hardly affects the results.
%

\subsection{Resolving Catastrophic Overfitting}
\label{ResolvingCatastrophicOverfitting}
First, we consider resolving the catastrophic overfitting in single-step AT. 
We set the Fast AT \cite{wong2020fast}, i.e., FGSM AT with a random initialization, as the baseline, and also consider other recently proposed methods for preventing the overfitting, including GradAlign (with a recommended coefficient $\lambda=0.2$) \cite{andriushchenko2020understanding}, Stable Single-AT (with $c=3$ check points) \cite{kim2020understanding}, Free AT ($m=8$) \cite{shafahi2019adversarial} and GAT \cite{sriramanan2020guided} (with a default coefficient $\lambda=10$).
We evaluate robustness using PGD-50 attack \cite{wong2020fast, andriushchenko2020understanding} and set the training epochs to 200 \cite{kim2020understanding} to closely examine whether the overfitting will occur.

The results of different methods on both CIFAR-10 and CIFAR-100 datasets are presented in \cref{tab:single-step},
where we apply Sub-AT to different base methods with the subspaces extracted from the corresponding training trajectory. The base methods use a piecewise schedule, where the overfitting mostly occurs.
We observe that under piecewise learning rates of 200 epochs, Fast AT and FGSM AT with GradAlign still meet serious catastrophic overfitting. 
A carefully designed cyclic learning rate schedule \cite{wong2020fast, andriushchenko2020understanding} helps overcome the overfitting, 
but it leads to an inferior robust performance (\eg, $-\mathbf{1.86}\%$ on CIFAR-10) compared with GradAlign Sub-AT.
Without any other regularization technique, our naive Fast Sub-AT is already able to obtain better robustness than FGSM AT with GradAlign regularization \cite{andriushchenko2020understanding}. 
GAT \cite{sriramanan2020guided} indeed overcomes the catastrophic overfitting and achieves impressive robustness among base methods,
but it still potentially suffers from overfitting problems, as the difference between the best and final is large.
Combined with Sub-AT, we are able to mitigate the overfitting and consistently improve the robustness.

To ensure the robustness improvements, we conduct additional evaluations via Auto-Attack.
On CIFAR-10, our GAT Sub-AT achieves ${46.33\pm0.37}\%$ robust accuracy (best) while base GAT achieves ${45.29\pm0.53}\%$ (best), and on CIFAR-100, our GradAlign Sub-AT achieves ${21.64\pm0.22}\%$ (best) while cyclic GradAlign achieves ${20.30\pm0.11\%}$ (best).
Thus via Sub-AT, we obtain state-of-the-art robustness performance on single-step AT, further reducing the robustness gap to multi-step AT.
Note that our good performance does not rely on the results obtained during the DLDR sampling, as both vanilla Fast AT and GradAlign meet serious overfitting during the training and only obtain a poor result far from satisfactory.

\begin{table*}[htbp]
 \centering
 \caption{Results of multi-step AT and Sub-AT on CIFAR-10/100 against \textbf{PGD-20} attack with $\ell_2$ and $\ell_\infty$ norm perturbations.} 
 \label{tab:multistepAT}
 {\footnotesize
 \begin{tabular}{c|ccc|ccc|ccc}
    \toprule
    \multirow{2}{*}{Dataset} & \multirow{2}{*}{Norm} & \multirow{2}{*}{Radius}  &\multirow{2}{*}{Settings} &\multicolumn{3}{c|}{Robust Accuracy} &\multicolumn{3}{c}{Natural Accuracy}\\
    \cline{5-7} \cline{8-10}
    & & & &Best &Final &Diff. &Best &Final &Diff.\\
    \midrule
    \multirow{4}{*}{CIFAR-10} &\multirow{2}{*}{$\ell_\infty $}
    &\multirow{2}{*}{$\epsilon=\frac{8}{255}$}  &AT &51.09 &42.92 &8.17 &80.50 &82.92 &-2.42\\
    & & &Sub-AT &\textbf{52.79} &\textbf{52.31} &\textbf{0.48} &80.46 &80.47 &-0.01\\
    \cline{2-10}
    &\multirow{2}{*}{$\ell_2 $}
    &\multirow{2}{*}{$\epsilon=\frac{128}{255}$}  &AT &67.73 &65.21 &2.52 &88.17 &88.82 &-2.42\\
    & & &Sub-AT &\textbf{69.14} &\textbf{69.01} &\textbf{0.13} &88.87 &88.84 &-0.01\\
    \midrule
    \multirow{4}{*}{CIFAR-100} &\multirow{2}{*}{$\ell_\infty $}
    &\multirow{2}{*}{$\epsilon=\frac{8}{255}$} &AT &26.80 &19.38 &7.42 &52.29 &53.27 &-0.98\\
    & & &Sub-AT &\textbf{27.50} &\textbf{27.02} &\textbf{0.48} &52.41 &52.18 &0.23\\
    \cline{2-10}
    &\multirow{2}{*}{$\ell_2 $}
    &\multirow{2}{*}{$\epsilon=\frac{128}{255}$} &AT &40.21 &34.98 &5.23 &61.98 &60.28 &1.70\\
    & & &Sub-AT &\textbf{41.48} &\textbf{41.00} &\textbf{0.48} &62.62 &63.13 &-0.51\\
    \bottomrule
 \end{tabular}
 }
\end{table*}

\subsection{Towards Larger Steps and Larger Radius}
After resolving the catastrophic overfitting, we demonstrate that Sub-AT overcomes the overfittings in training with a large step and radius, further improving the robustness performance.

\begin{figure}[htbp]
    \centering
    \includegraphics[width=0.8\linewidth]{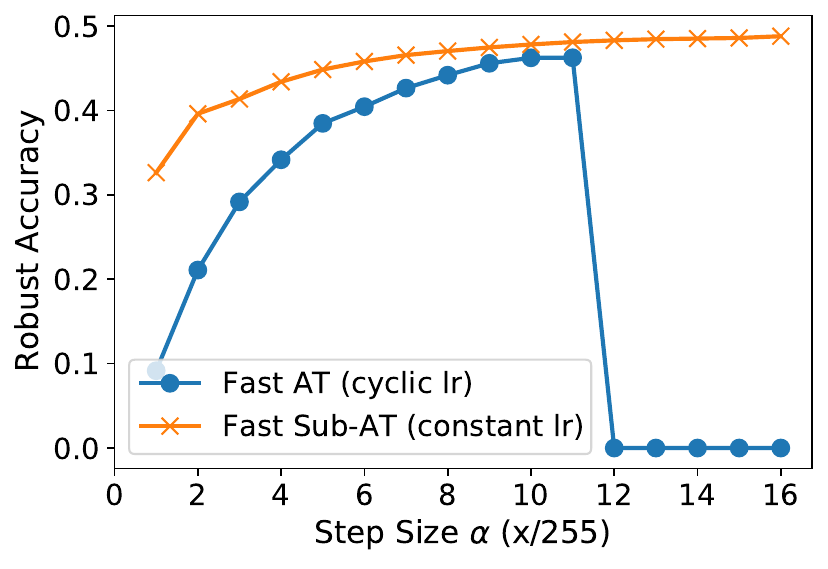}
    \caption{\textbf{Final} robust test accuracy of different single-step methods under PGD-20 attack  over different step sizes for $\epsilon=8/255$.}
    \label{fig:stepsize}
\end{figure}

\paragraph{Single-step AT with a larger step.}  
Although Wong \etal \cite{wong2020fast} discovered that adding a random initialization to FGSM AT could help avoid catastrophic overfitting, it only holds when the step size $\alpha$ is not too large. In fact, applying $\alpha$ larger than $12/255$ could still meet serious catastrophic overfitting (as illustrated in Fig.~3 of \cite{wong2020fast} for $\epsilon=8/255$). Since Sub-AT resolves the overfitting, we expect that it can allow training with a large $\alpha$. 
To this end, we consider CIFAR-10 and repeat Fast Sub-AT experiments in  \cref{ResolvingCatastrophicOverfitting} with $\alpha$ ranging from $1/255$ to $16/255$ and record the final robustness performance (note that we are in the same subspace).
From the results in \cref{fig:stepsize}, we observe that Sub-AT successfully resolves the overfitting for large step sizes and further improves the robustness, showing that using a large step indeed benefits robustness.

\begin{table*}[htbp]
 \centering
 \caption{
Results of applying Fast AT with a larger training radius $\epsilon$ to multi-step AT ($\ell_\infty$  attack of radius 8/255, CIFAR-10).
 }
 \label{tab:multi-step-subspace}
 {\footnotesize
 \begin{tabular}{l|c|ccc|ccc|c}
 \toprule
    \multirow{2}{*}{Method} &\multirow{2}{*}{Subspace} &\multicolumn{3}{c|}{Best} &\multicolumn{3}{c|}{Final} &\multirow{2}{*}{Time}\\
    \cline{3-5} \cline{6-8}
    & &Natural &PGD-50 &AA &Natural &PGD-50 &AA &\\
 \midrule
 PGD-10 AT &-- &80.50 &50.79 &47.29 &82.92 &42.51 &41.08 &7.0h\\
 PGD-10 Sub-AT &PGD-10 AT &80.46 &52.48 &48.37 &80.47 &52.01 &47.88  &4.9h \\
 Fast Sub-AT ($\epsilon=12/255$)&PGD-10 AT &81.02 &53.32 &48.58 &81.25 &52.85 &48.21 &3.9h\\
 Fast Sub-AT ($\epsilon=16/255$) &PGD-10 AT &79.63 &\textbf{54.17} &\textbf{49.14} &79.94 &\textbf{54.11} &\textbf{48.86} &3.9h \\
 \bottomrule
 \end{tabular}
 }
\end{table*}

\begin{figure}[htbp]
    \centering
    \includegraphics[width=0.8\linewidth]{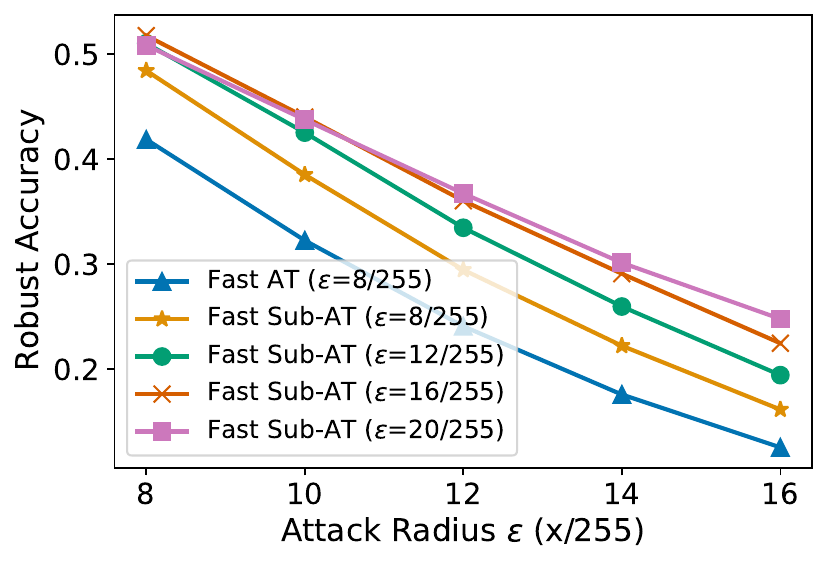}
    \caption{Test robust accuracy under PGD-20 attack with respect to different perturbation radius. We evaluate the checkpoint with the {best} performance on the validation set.}
    \label{fig:attack}
\end{figure}

\paragraph{Single-step AT with a larger radius.}
AT with a larger radius generally could provide better robustness guarantees against potential adversarial attacks. However, it is hard to train a model that is robust to a large $\epsilon$, especially for single-step AT \cite{andriushchenko2020understanding}. By constraining the training in subspace, we can conduct single-step AT for a large $\epsilon$ with ease. 
Similar to last section, we repeat the Fast Sub-AT experiments in  \cref{ResolvingCatastrophicOverfitting} with training radius $\epsilon \in [8/255, 20/255]$ (by default $\alpha=1.25\epsilon$). We select the best checkpoints (against PGD-20 attack of radius $8/255$) of different settings and plot their robust accuracy curves with respect to different strengths of attack.
In \cref{fig:attack}, we observe that within a certain range, training with a larger radius consistently improves robustness against attacks of different radii (especially for the large one), showing that the model's robustness is genuinely improved.
However, there also exist limitations as expected, as excessive perturbations will harm the valuable information of training data, resulting in a degenerated performance. 
For example, the best robustness against $8/255$ attack is achieved with training radius $16/255$.


We summarize the results of training with larger steps and larger radius in  \cref{tab:largestepandlargeradius}, where we also report total time consumption (including DLDR phase for Sub-AT) as well as evaluations on PGD-50 and Auto-Attack.
Main findings include:
\emph{\textbf{i}}) simply by increasing the training step size and radius, Sub-AT significantly improves model robustness against both strong attacks, without suffering catastrophic overfitting nor additional time cost;
\emph{\textbf{ii}}) notably, pure single-step-based Sub-AT achieves a very competitive robust accuracy under Auto-Attack than multi-step PGD-10 AT and even better robust accuracy under PGD-50 attack. We also get rid of the serious problem of robust overfitting, which PGD-10 AT suffers. 
\emph{\textbf{iii}}) more promisingly, Sub-AT only takes a seventh of total training time compared with PGD-10 AT, which is a considerable superiority.

\subsection{Extensions to Multi-step AT}
\label{MitigateRobustOverfitting}
We then apply Sub-AT to mitigate robust overfitting in multi-step AT, where we set standard PGD-10 AT \cite{madry2017towards} as the baseline.
To numerically show the degree of overfitting, we report the difference between the best and final robust accuracy.
The difference is crucial as, generally, we can only attain the final performance without using validation set.
From the results in \cref{tab:multistepAT}, we observe that robust overfitting occurs in every setting of AT as expected, and the gap between best and final can be as large as $8.17\%$ (on CIFAR-10 with $\ell_\infty$ norm).
By restricting AT in subspace, we reduce the gap to less than $0.5\%$ meanwhile significantly improving the robust accuracy (\eg, $+\mathbf{1.7}\%$ on CIFAR-10). 
In the good subspace extracted from DLDR \cite{li2021low}, generalization on clean data is also kept compared with standard AT.
More examinations via Auto-Attack are presented in \cref{multistep_autoattack},
where we observe that Sub-AT can indeed mitigate robust overfitting and consistently improve the robustness, 
rather than as a result of gradient masking \cite{papernot2017practical, athalye2018obfuscated}.
We note that the improvements in robustness naturally come from the low-dimensional optimization, and the results are promising to be further improved by combining other enhancement techniques, such as modifications on loss function \cite{chen2020robust}.

Finally, we apply Fast Sub-AT to the subspace extracted from multi-step PGD-10 AT. From the results in  \cref{tab:multi-step-subspace}, we observe that, surprisingly, 
Fast Sub-AT with a larger training radius remarkably outperforms the PGD-10 Sub-AT.
It implies that we can achieve strong robustness only with the guidance of weak adversarial examples in the subspace, which also brings computational benefits. This encouraging finding reveals the previously underestimated potential of single-step AT and can provide a new scheme for designing more robust and efficient AT methods.

\section{Conclusion}
In this paper, we focus on the serious catastrophic overfitting in single-step AT. 
From a novel perspective of optimization, we reveal the close link between the fast-growing gradient of each sample and overfitting, which can also explain the robust overfitting in multi-step AT.
To control the growth of gradient, Sub-AT is proposed to constrain AT in a carefully extracted subspace.
It successfully resolves both kinds of overfitting and hence significantly improves the robustness. 
Leveraging the subspace, we allow single-step AT with larger steps and radius, further improving the robustness. 
Weak adversarial examples generated from single-step AT can be trained to obtain even better robustness than those from multi-step PGD AT in subspace, revealing the great potential of single-step AT. 
As a result, pure single-step-based AT achieves comparable robustness to standard PGD-10 AT with only one-seventh of the training time, which is a solid step towards more efficient AT methods.

\section*{Acknowledgements}
We are very grateful for Hongkai Zheng, Qinghua Tao and anonymous reviewers
for the useful feedback on the paper.
The research leading to these results has received funding from National Key Research and Development Project
(No.2018AAA0100702), National Natural Science Foundation of China 61977046, and Shanghai Municipal Science and
Technology Major Project (2021SHZDZX0102). 
X. Huang is the corresponding author. 
\pagebreak

{\small
\bibliographystyle{ieee_fullname}
\bibliography{main}

\begin{thebibliography}{10}\itemsep=-1pt

\bibitem{andriushchenko2020understanding}
Maksym {Andriushchenko} and Nicolas {Flammarion}.
\newblock Understanding and improving fast adversarial training.
\newblock In {\em Proceedings of the Advances In Neural Information Processing
  Systems 33 (NeurIPS)}, volume~33, pages 16048--16059, 2020.

\bibitem{athalye2018obfuscated}
Anish Athalye, Nicholas Carlini, and David Wagner.
\newblock Obfuscated gradients give a false sense of security: Circumventing
  defenses to adversarial examples.
\newblock In {\em International Conference on Machine Learning (ICML)}, pages
  274--283. PMLR, 2018.

\bibitem{chen2020robust}
Tianlong Chen, Zhenyu Zhang, Sijia Liu, Shiyu Chang, and Zhangyang Wang.
\newblock Robust overfitting may be mitigated by properly learned smoothening.
\newblock In {\em International Conference on Learning Representations (ICLR)},
  2020.

\bibitem{cohen2019certified}
Jeremy Cohen, Elan Rosenfeld, and Zico Kolter.
\newblock Certified adversarial robustness via randomized smoothing.
\newblock In {\em International Conference on Machine Learning (ICML)}, pages
  1310--1320. PMLR, 2019.

\bibitem{croce2020robustbench}
Francesco {Croce}, Maksym {Andriushchenko}, Vikash {Sehwag}, Nicolas
  {Flammarion}, Mung {Chiang}, Prateek {Mittal}, and Matthias {Hein}.
\newblock Robustbench: a standardized adversarial robustness benchmark.
\newblock In {\em Advances in Neural Information Processing Systems (NeurIPS)
  Datasets and Benchmarks Track (Round 2)}, 2020.

\bibitem{croce2020reliable}
Francesco Croce and Matthias Hein.
\newblock Reliable evaluation of adversarial robustness with an ensemble of
  diverse parameter-free attacks.
\newblock In {\em International Conference on Machine Learning (ICML)}, pages
  2206--2216. PMLR, 2020.

\bibitem{deng2009imagenet}
Jia Deng, Wei Dong, Richard Socher, Li-Jia Li, Kai Li, and Li Fei-Fei.
\newblock Imagenet: A large-scale hierarchical image database.
\newblock In {\em the IEEE/CVF Conference on Computer Vision and Pattern
  Recognition (CVPR)}, pages 248--255, 2009.

\bibitem{dong2018boosting}
Yinpeng Dong, Fangzhou Liao, Tianyu Pang, Hang Su, Jun Zhu, Xiaolin Hu, and
  Jianguo Li.
\newblock Boosting adversarial attacks with momentum.
\newblock In {\em the IEEE/CVF Conference on Computer Vision and Pattern
  Recognition (CVPR)}, pages 9185--9193, 2018.

\bibitem{goodfellow2014explaining}
Ian~J Goodfellow, Jonathon Shlens, and Christian Szegedy.
\newblock Explaining and harnessing adversarial examples.
\newblock In {\em International Conference on Learning Representations (ICLR)},
  2015.

\bibitem{gressmann2020improving}
Frithjof {Gressmann}, Zach {Eaton-Rosen}, and Carlo {Luschi}.
\newblock Improving neural network training in low dimensional random bases.
\newblock In {\em Advances in Neural Information Processing Systems (NeurIPS)},
  volume~33, pages 12140--12150, 2020.

\bibitem{gur2018gradient}
Guy Gur-Ari, Daniel~A Roberts, and Ethan Dyer.
\newblock Gradient descent happens in a tiny subspace.
\newblock {\em arXiv preprint arXiv:1812.04754}, 2018.

\bibitem{hardt2016train}
Moritz Hardt, Ben Recht, and Yoram Singer.
\newblock Train faster, generalize better: Stability of stochastic gradient
  descent.
\newblock In {\em International Conference on Machine Learning (ICML)}, pages
  1225--1234. PMLR, 2016.

\bibitem{he2016deep}
Kaiming He, Xiangyu Zhang, Shaoqing Ren, and Jian Sun.
\newblock Deep residual learning for image recognition.
\newblock In {\em the IEEE/CVF Conference on Computer Vision and Pattern
  Recognition (CVPR)}, pages 770--778, 2016.

\bibitem{ioffe2015batch}
Sergey Ioffe and Christian Szegedy.
\newblock Batch normalization: Accelerating deep network training by reducing
  internal covariate shift.
\newblock In {\em International Conference on Machine Learning (ICML)}, pages
  448--456. PMLR, 2015.

\bibitem{jiang2019fantastic}
Yiding Jiang, Behnam Neyshabur, Hossein Mobahi, Dilip Krishnan, and Samy
  Bengio.
\newblock Fantastic generalization measures and where to find them.
\newblock In {\em International Conference on Learning Representations (ICLR)},
  2020.

\bibitem{kang2021understanding}
Peilin Kang and Seyed-Mohsen Moosavi-Dezfooli.
\newblock Understanding catastrophic overfitting in adversarial training.
\newblock {\em arXiv preprint arXiv:2105.02942}, 2021.

\bibitem{kim2020understanding}
Hoki {Kim}, Woojin {Lee}, and Jaewook {Lee}.
\newblock Understanding catastrophic overfitting in single-step adversarial
  training.
\newblock In {\em AAAI}, pages 8119--8127, 2020.

\bibitem{krizhevsky2009learning}
Alex Krizhevsky, Geoffrey Hinton, et~al.
\newblock Learning multiple layers of features from tiny images.
\newblock {\em Technical Report}, 2009.

\bibitem{li2018measuring}
Chunyuan {Li}, Heerad {Farkhoor}, Rosanne {Liu}, and Jason {Yosinski}.
\newblock Measuring the intrinsic dimension of objective landscapes.
\newblock In {\em International Conference on Learning Representations (ICLR)},
  2018.

\bibitem{li2021low}
Tao Li, Lei Tan, Qinghua Tao, Yipeng Liu, and Xiaolin Huang.
\newblock Low dimensional landscape hypothesis is true: {DNNs} can be trained
  in tiny subspaces.
\newblock {\em arXiv preprint arXiv:2103.11154}, 2021.

\bibitem{liao2018defense}
Fangzhou Liao, Ming Liang, Yinpeng Dong, Tianyu Pang, Xiaolin Hu, and Jun Zhu.
\newblock Defense against adversarial attacks using high-level representation
  guided denoiser.
\newblock In {\em the IEEE/CVF Conference on Computer Vision and Pattern
  Recognition (CVPR)}, 2018.

\bibitem{zhang2018defense}
Fangzhou Liao, Ming Liang, Yinpeng Dong, Tianyu Pang, Xiaolin Hu, and Jun Zhu.
\newblock Defense against adversarial attacks using high-level representation
  guided denoiser.
\newblock In {\em the IEEE/CVF Conference on Computer Vision and Pattern
  Recognition (CVPR)}, pages 1778--1787, 2018.

\bibitem{madry2017towards}
Aleksander Madry, Aleksandar Makelov, Ludwig Schmidt, Dimitris Tsipras, and
  Adrian Vladu.
\newblock Towards deep learning models resistant to adversarial attacks.
\newblock In {\em International Conference on Learning Representations (ICLR)},
  2018.

\bibitem{moosavi2019robustness}
Seyed-Mohsen Moosavi-Dezfooli, Alhussein Fawzi, Jonathan Uesato, and Pascal
  Frossard.
\newblock Robustness via curvature regularization, and vice versa.
\newblock In {\em the IEEE/CVF Conference on Computer Vision and Pattern
  Recognition (CVPR)}, pages 9078--9086, 2019.

\bibitem{neyshabur2017exploring}
Behnam Neyshabur, Srinadh Bhojanapalli, David McAllester, and Nati Srebro.
\newblock Exploring generalization in deep learning.
\newblock In {\em Advances in Neural Information Processing Systems (NeurIPS)},
  pages 5947--5956, 2017.

\bibitem{papernot2017practical}
Nicolas Papernot, Patrick McDaniel, Ian Goodfellow, Somesh Jha, Z~Berkay Celik,
  and Ananthram Swami.
\newblock Practical black-box attacks against machine learning.
\newblock In {\em Proceedings of the 2017 ACM on Asia Conference on Computer
  and Communications Security}, pages 506--519, 2017.

\bibitem{papernot2016distillation}
Nicolas Papernot, Patrick McDaniel, Xi Wu, Somesh Jha, and Ananthram Swami.
\newblock Distillation as a defense to adversarial perturbations against deep
  neural networks.
\newblock In {\em the IEEE Symposium on Security and Privacy (SP)}, pages
  582--597, 2016.

\bibitem{rice2020overfitting}
Leslie Rice, Eric Wong, and Zico Kolter.
\newblock Overfitting in adversarially robust deep learning.
\newblock In {\em International Conference on Machine Learning (ICML)}, pages
  8093--8104. PMLR, 2020.

\bibitem{shafahi2019adversarial}
Ali {Shafahi}, Mahyar {Najibi}, Mohammad~Amin {Ghiasi}, Zheng {Xu}, John
  {Dickerson}, Christoph {Studer}, Larry~S. {Davis}, Gavin {Taylor}, and Tom
  {Goldstein}.
\newblock Adversarial training for free.
\newblock In {\em Advances in Neural Information Processing Systems (NeurIPS)},
  volume~32, pages 3353--3364, 2019.

\bibitem{smith2017cyclical}
Leslie~N Smith.
\newblock Cyclical learning rates for training neural networks.
\newblock In {\em 2017 IEEE Winter Conference on Applications of Computer
  Vision (WACV)}, pages 464--472. IEEE, 2017.

\bibitem{SongKNEK18}
Yang Song, Taesup Kim, Sebastian Nowozin, Stefano Ermon, and Nate Kushman.
\newblock Pixeldefend: Leveraging generative models to understand and defend
  against adversarial examples.
\newblock In {\em International Conference on Learning Representations (ICLR)},
  2018.

\bibitem{sriramanan2020guided}
Gaurang Sriramanan, Sravanti Addepalli, Arya Baburaj, et~al.
\newblock Guided adversarial attack for evaluating and enhancing adversarial
  defenses.
\newblock In {\em Advances in Neural Information Processing Systems (NeurIPS)},
  volume~33, pages 20297--20308, 2020.

\bibitem{tramer2017ensemble}
Florian Tram{\`e}r, Alexey Kurakin, Nicolas Papernot, Ian Goodfellow, Dan
  Boneh, and Patrick McDaniel.
\newblock Ensemble adversarial training: Attacks and defenses.
\newblock In {\em International Conference on Learning Representations (ICLR)},
  2018.

\bibitem{tuddenham2020quasi}
Mark Tuddenham, Adam Pr{\"u}gel-Bennett, and Jonathan Hare.
\newblock Quasi-newton's method in the class gradient defined high-curvature
  subspace.
\newblock {\em arXiv preprint arXiv:2012.01938}, 2020.

\bibitem{vinyals2012krylov}
Oriol Vinyals and Daniel Povey.
\newblock Krylov subspace descent for deep learning.
\newblock In {\em Artificial Intelligence and Statistics (AISTATS)}, pages
  1261--1268. PMLR, 2012.

\bibitem{vivek2020single}
BS Vivek and R~Venkatesh Babu.
\newblock Single-step adversarial training with dropout scheduling.
\newblock In {\em the IEEE/CVF Conference on Computer Vision and Pattern
  Recognition (CVPR)}, pages 947--956, 2020.

\bibitem{vivek2019regularizer}
BS Vivek, Arya Baburaj, and R~Venkatesh Babu.
\newblock Regularizer to mitigate gradient masking effect during single-step
  adversarial training.
\newblock In {\em CVPR Workshops}, pages 66--73, 2019.

\bibitem{wang2019improving}
Yisen Wang, Difan Zou, Jinfeng Yi, James Bailey, Xingjun Ma, and Quanquan Gu.
\newblock Improving adversarial robustness requires revisiting misclassified
  examples.
\newblock In {\em International Conference on Learning Representations (ICLR)},
  2019.

\bibitem{wong2018provable}
Eric Wong and Zico Kolter.
\newblock Provable defenses against adversarial examples via the convex outer
  adversarial polytope.
\newblock In {\em International Conference on Machine Learning (ICML)}, pages
  5286--5295. PMLR, 2018.

\bibitem{wong2020fast}
Eric {Wong}, Leslie {Rice}, and J.~Zico {Kolter}.
\newblock Fast is better than free: Revisiting adversarial training.
\newblock In {\em International Conference on Learning Representations (ICLR)},
  2020.

\bibitem{wu2020adversarial}
Dongxian Wu, Shu-Tao Xia, and Yisen Wang.
\newblock Adversarial weight perturbation helps robust generalization.
\newblock In {\em Advances in Neural Information Processing Systems (NeurIPS)},
  volume~33, pages 2958--2969, 2020.

\bibitem{zagoruyko2016wide}
Sergey {Zagoruyko} and Nikos {Komodakis}.
\newblock Wide residual networks.
\newblock In {\em British Machine Vision Conference (BMVC)}, 2016.

\bibitem{zhang2019you}
Dinghuai Zhang, Tianyuan Zhang, Yiping Lu, Zhanxing Zhu, and Bin Dong.
\newblock You only propagate once: Painless adversarial training using maximal
  principle.
\newblock In {\em Advances in Neural Information Processing Systems (NeurIPS)},
  pages 227--238, 2019.

\bibitem{zhang2019theoretically}
Hongyang Zhang, Yaodong Yu, Jiantao Jiao, Eric~P. Xing, Laurent~El Ghaoui, and
  Michael~I. Jordan.
\newblock Theoretically principled trade-off between robustness and accuracy.
\newblock In {\em International Conference on Machine Learning (ICML)}, pages
  7472--7482. PMLR, 2019.

\bibitem{zhang2020attacks}
Jingfeng Zhang, Xilie Xu, Bo Han, Gang Niu, Lizhen Cui, Masashi Sugiyama, and
  Mohan Kankanhalli.
\newblock Attacks which do not kill training make adversarial learning
  stronger.
\newblock In {\em International Conference on Machine Learning (ICML)}, pages
  11278--11287. PMLR, 2020.

\end{thebibliography}
}

\appendix   
\setcounter{table}{0}   
\setcounter{figure}{0}
\renewcommand{\thetable}{A\arabic{table}}
\renewcommand{\thefigure}{A\arabic{figure}}

\twocolumn[
\begin{@twocolumnfalse}
\section*{\centering{Supplementary Material for \\ \emph{Subspace Adversarial Training}} \\~}
\end{@twocolumnfalse}
]

\section{Detailed Sampling Strategy}
\label{sec:samplingstragegy}
We present a detailed sampling strategy of model checkpoints for DLDR \cite{li2021low} in \cref{tab:samplingstragegy}, where we consider the times of uniform sampling in each epoch, the number of sampling epochs, the total number of samplings, and the dimension of the subspace extracted from the samplings.
The general sampling strategy is quite simple: we uniformly sample a few checkpoints in every training epoch before the overfitting occurs.
For single-step AT, the exact time when catastrophic overfitting occurs may be slightly different for different runs. Our sampling strategy is a conservative and robust one. 
We suggest sampling as more as model checkpoints before the overfitting occurs to estimate the subspace more accurately and to achieve better results. 
Note that the total number of samplings $t$ is small (around 200), and thus the computational overhead on corresponding decomposition (a  $t\times t$ matrix) is small.
This explains why the computational cost of the decomposition is negligible compared to the total computational cost.
A more delicate sampling strategy design may improve the performance and could be an interesting topic for future works.

\begin{figure}[h]
    \centering
    \includegraphics[width=0.8\linewidth]{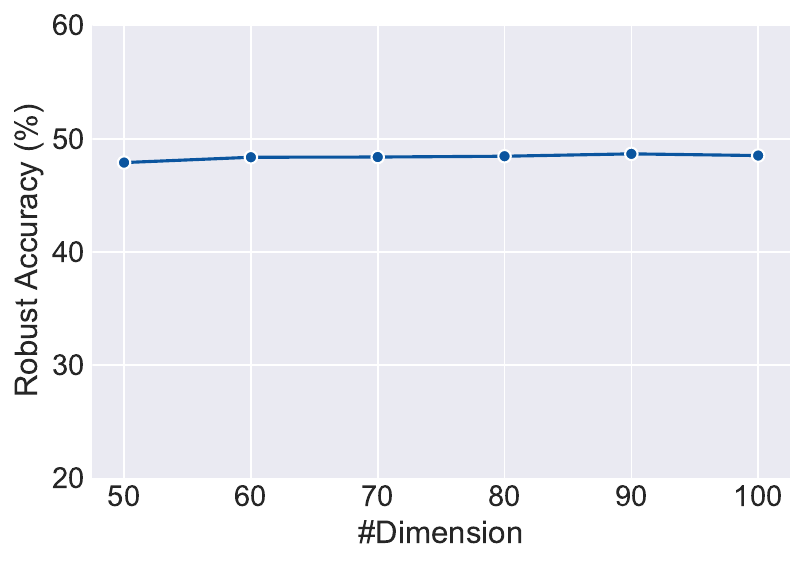}
    \caption{Ablation study for the dimension of subspace on CIFAR-10. Fast Sub-AT achieves similar best robustness performance across a wide range of dimensions. The robust accuracy is evaluated under PGD-20 attack.}
    \label{fig:components}
\end{figure}

\paragraph{Ablation study for the dimension $d$.}
We provide an ablation study for the dimension of subspace in \cref{fig:components}, where we vary the dimension of subspace for Fast Sub-AT from 50 to 100 and record the best robust accuracy obtained in corresponding subspaces.
We observe that the best robust accuracy varies very little across such a wide range of dimensions (\textless1\%). Thus we conclude that our Sub-AT is robust to the exact choice of the dimension $d$.

\section{Additional Results with Auto-Attack}
 \label{multistep_autoattack}
We further evaluate the performance of Sub-AT on mitigating robust overfitting in multi-step AT with Auto-Attack, a stronger and more reliable attack proposed by Croce \etal \cite{croce2020reliable}. 
From the results in \cref{tab:autoattack_cwattack}, we observe that indeed, Sub-AT effectively mitigates the robust overfitting and consistently improves the robust accuracy. Hence, training in subspace genuinely overcomes the overfitting and improves the model robustness rather than as a result of gradient masking \cite{papernot2017practical, athalye2018obfuscated}.


\begin{table}[htbp]
 \centering
 \caption{Robust accuracy of multi-step AT and Sub-AT on CIFAR-10/100 against $\ell_2$ and $\ell_\infty$ adversarial perturbations. The robust accuracy is evaluated under Auto-Attack \cite{croce2020reliable}. Our Sub-AT achieves consistent improvements on the robust accuracy while effectively mitigating the robust overfitting. The best robustness performances and the smallest difference between the best and the final are marked in \textbf{bold}.}
 \label{tab:autoattack_cwattack}
 { \small
 \begin{tabular}{c|cc|ccc}
    \toprule
    \multirow{2}{*}{Dataset} & \multirow{2}{*}{Norm}  &\multirow{2}{*}{Settings} &\multicolumn{3}{c}{Auto-Attack}\\
    \cline{4-6} 
    &  & &Best &Final &Diff. \\
    \midrule
    \multirow{4}{*}{CIFAR-10} &\multirow{2}{*}{$\ell_\infty$}
      &AT &47.29 &41.08 &6.21\\
    &  &Sub-AT &\textbf{48.37} &\textbf{47.88} &\textbf{0.49}\\
    \cline{2-6}
    &\multirow{2}{*}{$\ell_2$}
      &AT &65.66 &63.93 &1.73\\
    &  &Sub-AT &\textbf{66.99} &\textbf{67.15} &\textbf{-0.16}\\
    \midrule
    \multirow{4}{*}{CIFAR-100} &\multirow{2}{*}{$\ell_\infty$}
     &AT &22.83 &18.12 &4.71\\
    & &Sub-AT &\textbf{23.89} &\textbf{23.83} &\textbf{0.06}\\
    \cline{2-6}
    &\multirow{2}{*}{$\ell_2$}
     &AT &36.91 &32.70 &4.21\\
    & &Sub-AT &\textbf{38.14} &\textbf{37.84} &\textbf{0.30}\\
    \bottomrule
 \end{tabular}
 }
\end{table}

\section{Results on Tiny-ImageNet}
\label{TinyImageNet}
The results on Tiny-ImageNet \cite{deng2009imagenet} is presented in \cref{tab:tinyimagenet},
where we use PreAct ResNet18 model \cite{he2016deep} and train it for 100 epochs with learning rate decay at 50 and 80 following \cite{chen2020robust}. Sub-AT is trained for 20 epochs with a constant learning rate 1, accordingly. For single-step Fast AT, we observe that the catastrophic overfitting occurs at the 17th epoch, and thus we only sample 16 epochs for DLDR. Even with such few samplings, we obtain an 18.79\% single-step robust accuracy, while standard PGD-10 AT achieves a slightly better 19.84\% robust accuracy, but with around \textbf{12x} training time overhead and, even worse, serious robust overfitting problem. Within the better subspace extracted from PGD-10 AT, our Fast Sub-AT achieves \textbf{21.87}\% robust accuracy, which is significantly better than base PGD-10 AT and also enjoys computational benefits.

\begin{table*}[htbp]
 \centering
 \caption{Detailed sampling strategy for DLDR. We report the times of uniformly sampling in each epoch of training, the number of sampling epochs, the total number of samplings $t$, and the dimension of the subspace extracted from the samplings of parameters.
 Note that there is an additional sampling on the parameter initialization as we start the Sub-AT from the initialization. }
 \label{tab:samplingstragegy}
 \begin{tabular}{@{}c|c|ccccc@{}}
    \toprule
    Type &Datasets &Method &\#Times/{\small epoch} &\#Epochs &$t$ &\#Dimension \\
    \midrule
    \multirow{5}{*}{Single-step} 
    &\multirow{1}{*}{CIFAR-10} &Fast AT &2 &65 &131 &80\\
    &CIFAR-10 &GradAlign &2 &65 &131 &100 \\
    &CIFAR-10 &GAT &2 &100 &201 &100 \\
    &\multirow{1}{*}{CIFAR-100} &Fast AT &2 &65 &131 &80\\
    &CIFAR-100 &GradAlign &2 &65 &131 &100 \\
    &CIFAR-100 &GAT &1 &140 &141 &100 \\
    &\multirow{1}{*}{Tiny-ImageNet} &Fast AT &4 &16 &65 &50\\
    \midrule
    \multirow{3}{*}{Multi-step} &\multirow{1}{*}{CIFAR-10} &PGD-10 AT &2 &100 &201 &120\\
    &\multirow{1}{*}{CIFAR-100} &PGD-10 AT &2 &100 &201 &120\\
    &\multirow{1}{*}{Tiny-ImageNet} &PGD-10 AT &4 &50 &201 &120\\
    \bottomrule
 \end{tabular}
\end{table*}

\begin{table*}[htbp]
 \centering
 \caption{Results on Tiny-ImageNet. We use PreAct ResNet18 model and consider both single-step and multi-step AT. Sub-AT demonstrates its superior performance in both robustness performance and computational overhead. The time consumption is evaluated on an Nvidia Tesla V100.
 }
 \label{tab:tinyimagenet}
 {
 \begin{tabular}{ll|c|cc|cc|r}
 \toprule
    &\multirow{2}{*}{Method} &\multirow{2}{*}{Subspace} &\multicolumn{2}{c|}{Best} &\multicolumn{2}{c|}{Final} &\multirow{2}{*}{Time}\\
    \cline{4-5} \cline{6-7}
    & & &Natural &PGD-20  &Natural &PGD-20  &\\
 \midrule
 \textbf{Single-step} &Fast AT &-- &28.79 &11.90 &42.54 &0.00 &3.5h\\
  &Fast Sub-AT ($\epsilon=8/255$) &Fast AT &39.38 &16.75 &39.19 &16.17 &1.3h\\
  &Fast Sub-AT ($\epsilon=12/255$) &Fast AT &38.11 &18.52 &38.306 &18.13 &1.3h\\
  &Fast Sub-AT ($\epsilon=16/255$) &Fast AT &37.32 &\textbf{18.79} &37.20 &\textbf{18.22} &1.3h\\
 \midrule
 \textbf{Multi-step} 
 &PGD-10 AT &--  &42.76 &19.84 &46.57 &14.18 &15.7h\\
    &PGD-10 Sub-AT &PGD-10 AT &40.82 &20.52    &40.78 &19.55 &12.1h\\
     &PGD-10 AT ($\epsilon=12/255$) &PGD-10  AT &40.88 &21.42 &42.27 &21.41 &8.6h\\
 &PGD-10 AT ($\epsilon=16/255$) &PGD-10 AT &40.99 &\textbf{21.87}    &41.27 &\textbf{21.60} &8.6h\\
 \bottomrule
 \end{tabular}
 }
\end{table*}
\begin{table*}[!h]
 \caption{ Results on multi-step PGD-10 AT with WideResNet-28-10 model against PGD-20 attack ($\ell_\infty$ norm, $\epsilon=8/255$). }
 \label{tab:Wide-ResNet}
 \centering
{
 \begin{tabular}{cc|ccc|ccc}
    \toprule
    \multirow{2}{*}{Dataset}  &\multirow{2}{*}{Settings} &\multicolumn{3}{c|}{Robust Accuracy} &\multicolumn{3}{c}{Natural Accuracy}\\
     & &Best &Final &Diff. &Best &Final &Diff.\\
    \midrule
    \multirow{2}{*}{CIFAR-10} &AT &53.26 &46.70 &6.56 &84.69 &{85.81} &-1.12\\
   &Sub-AT &\textbf{55.14} &\textbf{54.75} &\textbf{0.39} &{84.79} &84.71 &0.08\\
   \midrule
  \multirow{2}{*}{CIFAR-100} &AT &29.44 &23.26 &6.18 &{57.64} &56.08 &1.56\\
   &Sub-AT &\textbf{31.07} &\textbf{30.69} &\textbf{0.38} &57.24 &{57.40} &-0.16\\
    \bottomrule
 \end{tabular}
 }
\end{table*}

\section{Results on Wide-ResNet}
\label{WideResNet}
We conduct further experiments on Wide-ResNet \cite{zagoruyko2016wide} architectures. 
Specifically, we use the WideResNet-28-10 model and consider $\ell_\infty$ adversarial perturbations with radius $8/255$ on CIFAR-10 and CIFAR-100. 
From the results in \cref{tab:Wide-ResNet}, we observe that similarly, robust overfitting can be successfully mitigated by Sub-AT, meanwhile with significant improvements in robustness. We achieve +\textbf{1.88}\% on CIFAR-10 and +\textbf{1.63}\% on CIFAR-100, and successfully control the robust accuracy gap (between the best and the final) under 0.4\%.
Thus we conclude that Sub-AT can be easily applied to other architectures and obtain similar enhancements in robustness.

\end{document}